\documentclass[10pt,twocolumn,letterpaper]{article}

\usepackage[pagenumbers]{iccv} 

\usepackage{amsmath,amssymb}

\usepackage{booktabs}
\usepackage{multirow}
\usepackage{multicol}
\usepackage[mathscr]{euscript}
\usepackage[dvipsnames]{xcolor}
\usepackage{colortbl}
\usepackage{makecell}
\usepackage{subcaption}
\usepackage[symbol]{footmisc}

\definecolor{shadecolor}{rgb}{0.92, 0.92, 0.92}
\definecolor{gtgray}{gray}{0.97}
\definecolor{mygray}{gray}{.88}
\definecolor{gray1}{gray}{.90}
\definecolor{gray2}{gray}{.92}
\definecolor{gray3}{gray}{.94}

\definecolor{iccvblue}{rgb}{0.21,0.49,0.74}
\usepackage[pagebackref,breaklinks,colorlinks,allcolors=iccvblue]{hyperref}

\def\paperID{2366} 
\def\confName{ICCV}
\def\confYear{2025}

\title{GestureLSM: Latent Shortcut based Co-Speech Gesture Generation with Spatial-Temporal Modeling}

\author{Pinxin Liu$^{1}$, Luchuan Song$^{1}$\footnotemark[1], Junhua Huang$^{1}$, Haiyang Liu$^{2}$, Junfan Zhu$^{3}$, Chenliang Xu$^{1}$\\
$^{1}$University of Rochester, $^{2}$University of Tokyo, $^{3}$University of Chicago}

\begin{document}

\twocolumn[{%
\renewcommand\twocolumn[1][]{#1}%
\maketitle
\begin{center}
    \centering
    \vspace{-0.7cm}
    \captionsetup{type=figure}
    \includegraphics[width=1\textwidth]{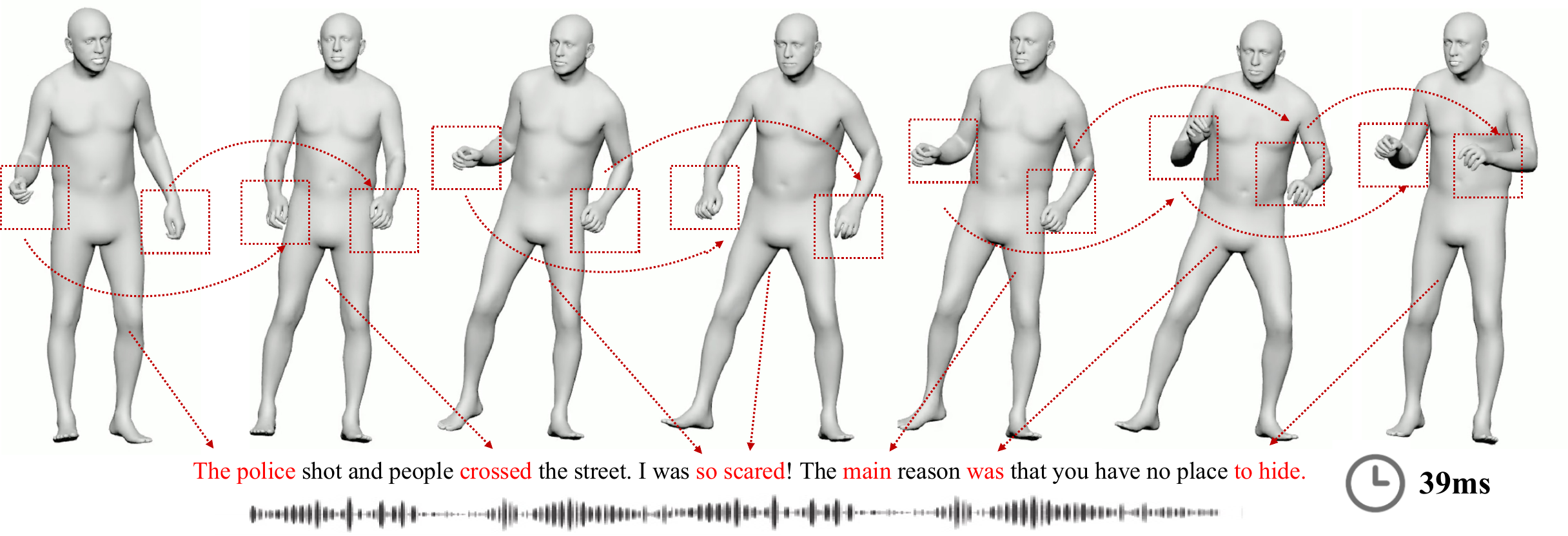}
    \vspace{-0.3cm}
    \captionof{figure}{\textbf{GestureLSM.} We present Gesture Latent Shortcut Model, a method that generates full-body human gestures from speech and text scripts with high quality and real-time speed. It explicitly models the body regions interactions, \textit{e.g.}, the interactions between body and hands, to achieve coherent gesture motions.  Besides, it is also capable of real-time generation based on shortcut sampling.}
\label{Teaser}
\end{center}%
}]
\footnotetext[1]{corresponding author.}

\begin{abstract}
    Generating full-body human gestures based on speech signals remains challenges on quality and speed. Existing approaches model different body regions such as body, legs and hands separately, which fail to capture the spatial interactions between them and result in unnatural and disjointed movements. Additionally, their autoregressive/diffusion-based pipelines show slow generation speed due to dozens of inference steps. To address these two challenges, we propose \textbf{GestureLSM}, a flow-matching-based approach for Co-Speech Gesture Generation with spatial-temporal modeling. Our method i) explicitly model the interaction of tokenized body regions through spatial and temporal attention, for generating coherent full-body gestures. ii) introduce the flow matching to enable more efficient sampling by explicitly modeling the latent velocity space. To overcome the suboptimal performance of flow matching baseline, we propose latent shortcut learning and beta distribution time stamp sampling during training to enhance gesture synthesis quality and accelerate inference. 
    Combining the spatial-temporal modeling and improved flow matching-based framework, GestureLSM achieves state-of-the-art performance on BEAT2 while significantly reducing inference time compared to existing methods, highlighting its potential for enhancing digital humans and embodied agents in real-world applications. Project Page: {\small\url{https://andypinxinliu.github.io/GestureLSM}}
\end{abstract}

\section{Introduction}
\label{sec:intro}

Speech is often accompanied by gestures that serve to convey emotions and enhance comprehension~\cite{de2012interplay,song2023emotional} within daily conversations. These non-verbal cues play a crucial role in effective communication~\cite{burgoon1990nonverbal}, with gesture generation serving as a fundamental component of human-computer interaction. As artificial intelligence continues to advance, endowing virtual avatars with realistic gestural capabilities will be essential for fostering immersive interactive experiences and digital avatar construction~\cite{tang2025generative, adaptive, huang2024modelingdrivinghumanbody,song2024texttoon,song2024tri,song2021talking}.


Many recent works~\cite{mambatalk, liu2025contextualgesturecospeechgesture,liu2023emage,yi2022generating} have conducted preliminary studies on full-body co-speech gesture generation. These methods typically treat different body regions as separated motion features, represented using either continuous~\cite{habibie2021learning} or discrete \textit{VQ} tokens~\cite{liu2023emage,yi2022generating}. While these approaches have demonstrated plausible gesture patterns for individual body regions, they often produce coherent full-body movements. This issue stems from their separated motion features learning process, which overlooks the nuanced interactions between different body parts. 


For example, when expressing the sentence \textit{``I completely agree"}, natural gestures involve intricate coordination. The fingers may point or emphasize, the arms may extend outward, and the torso may subtly shift, all contributing to the reinforcement of affirmation.
However, separated motion decoding constraints models from learning such fine-grained interactions, leading to uncoordinated and unnatural gesture patterns. These limitations not only hinder the realism of generated gestures but also restrict their applications in industry-grade animations.

Moreover, many existing models~\cite{diffsheg,chen2024syntalker} employ either diffusion-based approaches or auto-regressive generation methods~\cite{yi2022generating}. The diffusion-based approaches operate through iterative denoising processes, whereas auto-regressive models generate gestures sequentially. Both approaches introduce substantial computational overhead, thereby hindering real-time inference and posing significant challenges for real-world deployment.

To address the above issues, we propose \textbf{GestureLSM}: a flow-matching-based approach for Co-Speech Gesture Generation with spatial-temporal modeling.. Our method explicitly models the interactions between different body parts by segmenting the human body into the upper body, hands, lower body, and facial expressions. To fully capture spatial-temporal dependencies, we first apply spatial attention to ensure coherence between body regions at each time step, followed by temporal attention to model motion progression. This sequential design enables our model to learn both intra-frame structural relationships and inter-frame motion dynamics.
In addition, we introduce the flow matching to enable more efficient sampling by explicitly modeling the latent velocity space. 
Even though flow matching is potentially faster and easier for learning in theory, we found the vanilla flow matching yields suboptimal results' quality and speed, \textit{ i.e.}, underperform the diffusion-based models. To solve this, we adopt latent shortcut learning and beta distribution time stamp sampling during training to enhance gesture synthesis quality and accelerate inference, which are the keys to enable flow matching based model outperform diffusion based model on our task. Combining the spatial-temporal modeling and improved flow matching framework, GestureLSM achieves state-of-the-art performance on BEAT2 while significantly reducing inference time compared to existing methods.
In summary, our primary contributions are: 
\begin{itemize}
    \item We present the \textit{GestureLSM}, a framework that achieves high quality and real-time co-speech human gesture generation.
    \item We develop the \textit{spatial-temporal modeling} that enables cohernet gesture generation by capturing and integrating interactions across different body regions.
    \item We improve the flow matching-based to outperform diffusion-based models for gesture generation, in particular, we introduce the latent shortcut model and optimize time-sampling strategies during training to accelerate inference speed and benefit the gesture synthesis.
\end{itemize}

\begin{figure}[]
\centering
  \includegraphics[width=.9\columnwidth, trim={0cm 0cm 0cm 0cm}, clip]{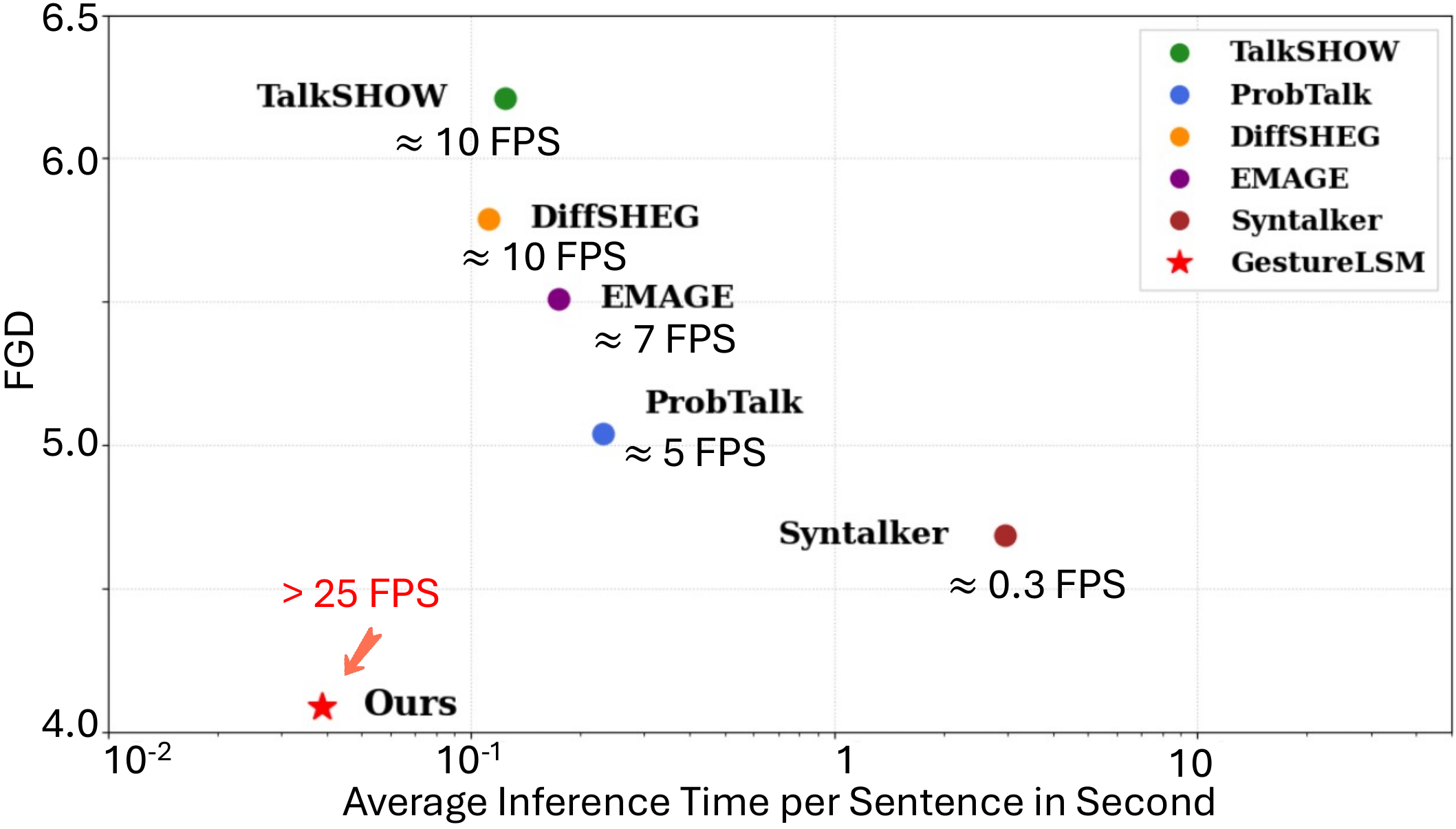}
\vspace{-0.3cm}
\caption{Our GestureLSM achieves significant generation quality improvement over baseline methods with fastest inference speed. The inference time is computed on one NVIDIA A100 while the generation quality is from FGD on BEAT2.   
}

\vspace{-3mm}
\label{fig:speed}
\end{figure}
\section{Related Works}

\paragraph{Co-speech Gesture Generation}
Existing works on co-speech gesture generation mostly employ skeleton- or joint-level pose representations.
\cite{ginosar2019gestures} use an adversarial framework to predict hand and arm poses from audio, and leverage conditional generation~\cite{EverybodyDanceNow} based on pix2pixHD~\cite{wang2018pix2pixHD} for videos.
Some recent works~\cite{liu2022learning,Deichler_2023,xu2023chaingenerationmultimodalgesture, liu2024tango, zhang2024kinmokinematicawarehumanmotion, liu2025contextualgesturecospeechgesture,song2021fsft,liu2025intentionalgesturedeliverintentions,liu2025semgessemanticsawarecospeechgesture} learn the hierarchical semantics or leverage contrastive learning to obtain joint audio-gesture embeding to assist the gesture pose generation. HA2G~\cite{liu2022learning} construct high and low level audio-motion embedding for gesture generation. TalkShow~\cite{yi2022generating} estimates SMPL~\cite{SMPL-X:2019} poses, and models the body and hand motions for talk-show videos. CaMN~\cite{liu2022beat} and EMAGE~\cite{liu2023emage} propose a large scale conversational and speech datasets for joint face and body modeling with diverse style control, with GPT-style decoding for gesture generation. 

MambaTalk~\cite{mambatalk} speeds up the generation process with an efficient mamba structure. Semantic Gesticulator~\cite{semanticgesticulator} enhances gesture generation with semantic annotation of gesture types and triggers for retrieval. DiffSHEG~\cite{diffsheg} and SynTalker~\cite{chen2024syntalker} build up a diffusion-based gesture generation pipeline. However, none of these works considers the interactions of different body regions conditioned on speech input and achieves fast and real-time gesture generations.

\vspace{-0.1cm}
\paragraph{Fast Diffusion Sampling}
Diffusion models~\cite{ho2020denoisingdiffusionprobabilisticmodels,Lu_2023_ICCV,li2025setstraightautosteeringdenoising,Lu_2024_CVPR,lu2024robust} have demonstrated impressive generation quality across various modalities~\cite{ruan2022mmdiffusion,stablevideodiffusion,ning2024dctdiff}, but suffer from slow inference speeds due to their iterative sampling process. To address these efficiency challenges, several approaches have been proposed, such as Consistency Models~\cite{song2023consistency,luo2023latent} and Diffusion Distillations~\cite{sauer2024fast,yin2024one}. However, these methods still face limitations in terms of training speed and flexibility.

One promising direction for improving inference speed is the use of flow matching. \cite{liu2022flowstraightfastlearning} introduces rectified flow, which provides a crucial baseline for diffusion acceleration. Through rectification, they straighten the ODE path of flow-matching-based diffusion models. Building on this, \cite{liu2023instaflow,zhu2024slimflowtrainingsmalleronestep,wang2024rectifieddiffusionstraightnessneed,gao2025eraseanything,zhu2024oftsronestepflowimage} scales rectified flow to large-scale text-to-image generation, achieving one-step generation.

Shortcut Models~\cite{shortcutmodels} further unified the distillation with flow matching. It considers the distance for the future curvature and allow the model to predict a shortcut along the sampling trajectory through self-consistency. Based on this, we further analyze the denoising patterns and explore the time distribution sampling to enhance the generation quality and inference speed for co-speech gesture generation.

\begin{figure*}[]
\centering
  \includegraphics[width=2\columnwidth, trim={0cm 0cm 0cm 0cm}, clip]{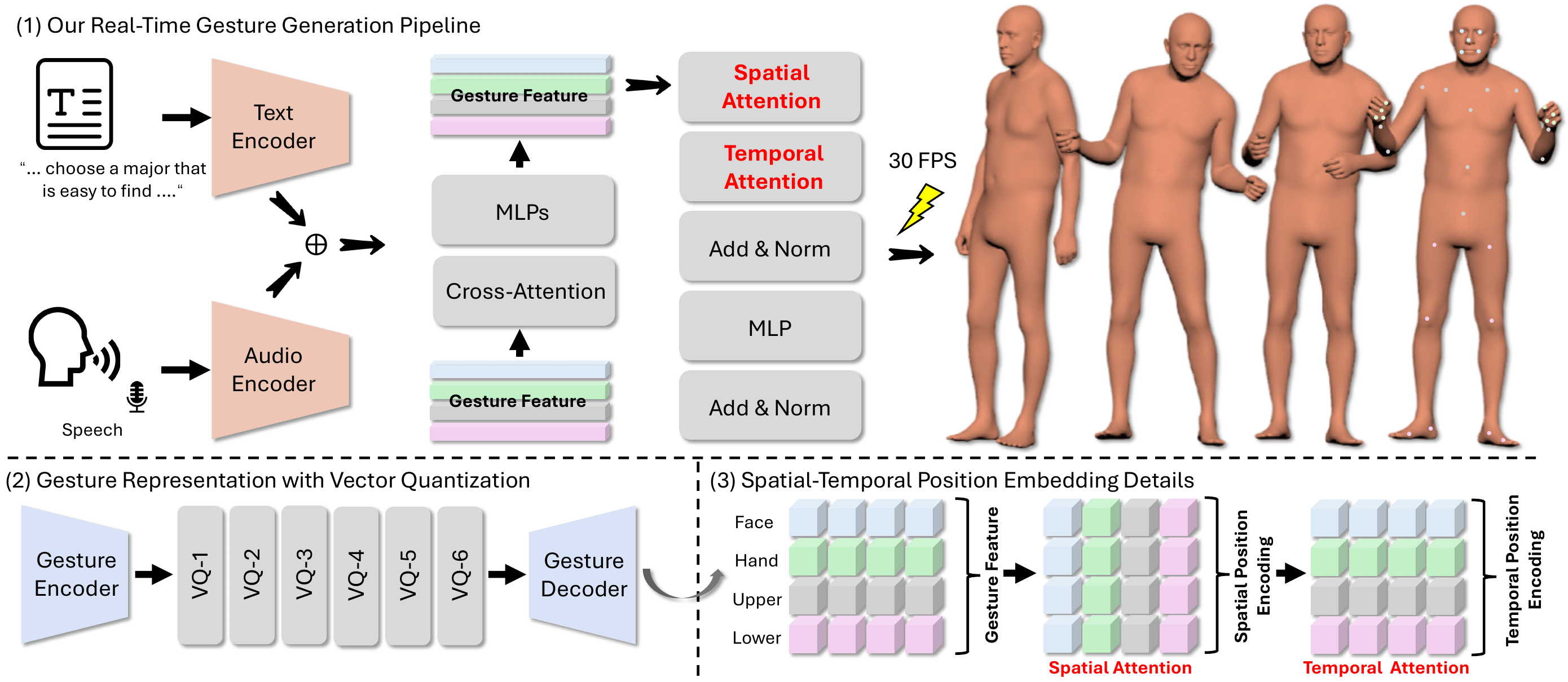}
  \vspace{-0.3cm}
\caption{The pipeline of GesutureLSM. (1) Our GestureLSM generate full-body gestures from speech and text scripts. The concatenated audio and text features are fused into gesture features via cross-attention. The condition fused gesture features are adopted to decode gesture latents with our proposed spatial-temporal decoder. The optimization objective is based on the flow matching (as shown in Figure \ref{fig:pipeline-2} for details.) (2) The gesture latents are from pretrained \textit{RVQ} (Residual Vector Quantization) models. (3) The details of spatial-temporal attention, which integrates with position encoding to learn the interaction of body regions.}
\label{fig:pipeline}
\vspace{-3mm}
\end{figure*}

\section{Method}
As illustrated in Fig.\ref{fig:pipeline}, our method achieves fast co-speech gesture generation with fine-grained spatial-temporal learning. We first construct the gesture motion representation as different body joint groups in Sec.~\ref{sec:quantization}. To achieve the learning of interactions of different body joints, we leverage spatial and temporal attention in Sec.~\ref{sec:spatial-temporal}. We achieve real-time, high quality gesture generations through shortcut modeling in Sec.~\ref{sec:rectified}

\subsection{Speech-Gesture Feature Representation}
\label{sec:quantization}

\paragraph{Gesture Body Region Quantization.}
We construct \textit{RVQ} quantizers~\cite{rvq} to convert the local body motions, namely hands, upper body, and leg motions, into discrete tokens, as illustrated on the bottom left of Fig.~\ref{fig:pipeline}.

For each body joint group with sequence length $T$, we employ a convolutional network to encode the body region $\mathbf{B}=\{\mathbf{b}_t\}_{t=1, ..., T}$ into vectors $\mathbf{V}=\{\mathbf{v}_t\}_{t=1, ..., T}$, where each vector $\mathbf{v}_t$ corresponds to time $t$, by the encoder network $\mathcal{E}$, composed of two convolutional residual blocks.
Then we quantize the vector with a codebook $\mathbf{C}=\{\mathbf{c}_i\}_{i=1}^{C}$ by replacing the vector $\mathbf{v}_t$ with its nearest code entry $\mathbf{\tilde{v}}_t$ as
\[
\begin{cases}
\mathbf{\tilde{v}}_t = \mathcal{Q}(\mathbf{v}_t), \\
\mathcal{Q}(\mathbf{v}_t) = \mathbf{c}_i, 
\end{cases} i = \arg\min_{i} \|\mathbf{c}_i - \mathbf{v}_t\|_2.
\]
Here $\mathcal{Q}$ denotes the quantization process.
After the quantization, the decoder $\mathcal{D}$ decodes the approximate vectors $\mathbf{\tilde{v}}_t$ to get the original joint information, as: 
\begin{equation}
    \{ \mathbf{\tilde{b}}_t \} = \mathcal{D}~ ( \{ \mathbf{\tilde{v}}_t \} )
    \equiv \mathcal{D}~ (~ \mathcal{Q}~ (~ \mathcal{E}~ (~ \{ \mathbf{b}_t \} ~ ) ~ ) ~ ).
\end{equation}

\paragraph{Speech Signal Processing.} Following recent works~\cite{liu2023emage,chen2024syntalker}, we represent speech signals with both low-level onset information, characterized by amplitude and high-level semantics extracted from transcripts using BERT~\cite{devlin2018bert}. To process these components separately, we employ dedicated audio and text encoders, subsequently performing element-wise concatenation to integrate the two feature types. The fused speech representation is then incorporated into the gesture representation through multiple layers of cross-attention, where the gesture features act as queries, while the speech features serve as both keys and values.


\subsection{Spatial-Temporal Gesture Generation}  
\label{sec:spatial-temporal}  

We employ a transformer-based model to generate body gestures with audio conditions. Different from the prior approaches~\cite{liu2023emage,chen2024syntalker,yi2022generating}, which treat the entire body separately, our method explicitly models the interactions between different gesture groups. Specifically, we introduce two distinct attention mechanisms within the transformer: \textbf{spatial attention} and \textbf{temporal attention}, as illustrated in Fig.~\ref{fig:pipeline}. Our model first applies spatial attention to capture inter-region relationships within one frame, followed by temporal attention to model motion dynamics over time. We will describe these mechanisms and their strengths, and their limitations in below.  

\paragraph{Spatial Attention.}  
Spatial attention operates across body regions within a single frame, ensuring structural coherence between different body parts while disregarding temporal dependencies. As illustrated in Fig.~\ref{fig:pipeline}, we rearrange the input tokens such that all time steps are treated as independent batches, allowing attention to be computed only between different body regions at a given moment. The spatial attention mechanism is formulated as:  

\begin{equation}
    \mathcal{A}_{s} = \text{SoftMax}(\frac{Q_s K_s}{\sqrt{d}} + \mathbf{P}) V_s,
\end{equation}  
where \( Q_s, K_s, V_s \in \mathbb{R}^{n\times d} \), with \( n \) being the number of body regions and \( d \) the feature dimensionality. Before applying spatial attention, we incorporate \textbf{spatial positional encodings} to capture the relative positions of different body regions. This mechanism ensures consistency across different body parts in a single frame. However, it does not explicitly model how gestures evolve over time, which is crucial for generating natural motion.  

\paragraph{Temporal Attention.}  
Temporal attention, in contrast, operates along the time axis, capturing the motion dynamics of each body region independently. To achieve this, we rearrange the input tokens such that all body regions are treated as separate batches, allowing attention to be computed only across different time steps for each region. The temporal attention mechanism is defined as: 
\begin{equation}
    \mathcal{A}_{t} = \text{SoftMax}(\frac{Q_t K_t}{\sqrt{d}} + \mathbf{P}) V_t,
\end{equation}  
where \( Q_t, K_t, V_t \in \mathbb{R}^{T\times d} \), with \( T \) being the number of time steps. \textbf{Temporal positional encodings} are added before applying attention to ensure that the model learns the sequential nature of motion. This mechanism effectively captures gesture continuity and motion patterns across time. However, it does not consider interactions between different body regions at a given time step, which may lead to inconsistencies in complex gestures involving coordinated limb movements.  

\paragraph{Gesture Generator.}  
To fully capture spatial-temporal dependencies, we first apply spatial attention to ensure coherence between body regions at each time step, followed by temporal attention to model motion progression. This sequential design enables our model to learn both intra-frame structural relationships and inter-frame motion dynamics.  

Finally, we employ the feed-forward network (FFN) as in the standard Transformer~\cite{vaswani2023attentionneed} to further refine the attention features in each generator block.  

\begin{figure}[]
\centering
  \includegraphics[width=1\columnwidth, trim={0cm 0cm 0cm 0cm}, clip]{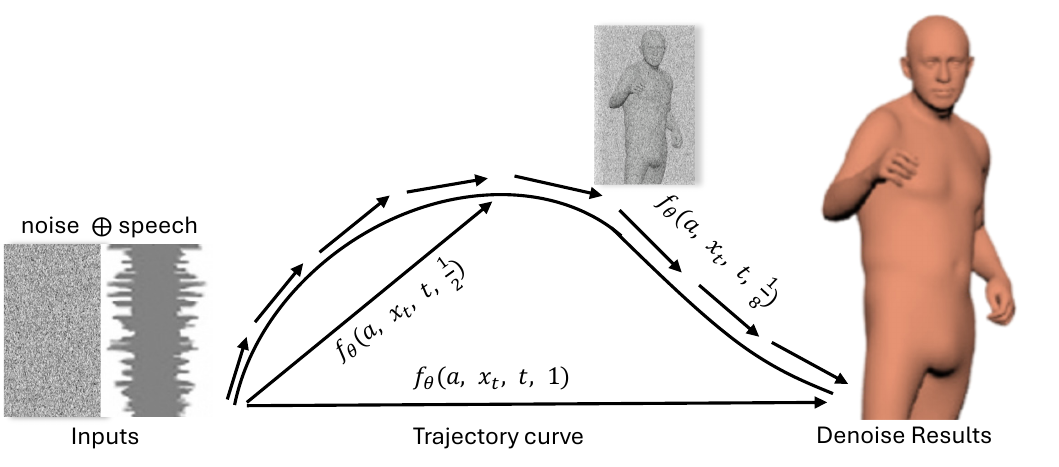}
\vspace{-3mm}
\vspace{-0.3cm}
\caption{The details of Latent Shortcut Model. GestureLSM enforces self-consistency in motion trajectories by ensuring that the overall movement predicted over a longer duration remains consistent with the sum of shorter movements. It encourages the model to maintain coherence in gesture generation, preventing drift and enhancing stability across different time steps.}
\label{fig:pipeline-2}
\end{figure}

\subsection{Gesture Latent Shortcut Model}
\label{sec:rectified}

\paragraph{Flow-matching and Shortcut Models.}
Recent flow-matching models~\cite{lipman2022flow,liu2022flowstraightfastlearning} tackle generative modeling by learning an ordinary differential equation (ODE) to transform noise into data. The model defines \( x_t \) as a linear interpolation between a data point \( x_1 \sim \mathcal{D} \) and a noise point \( x_0 \sim \mathcal{N}(0,\mathbb{I}) \), where:
%
\[
\begin{cases}
x_t = (1-t)\,x_0 + t\,x_1, \\
v_t = x_1 - x_0.
\end{cases}
\]
Given \( x_0 \) and \( x_1 \), the velocity \( v_t \) is fixed. However, for a given \( x_t \), there are multiple plausible pairs \( (x_0, x_1) \), leading to a distribution over possible velocities. Thus, \( v_t \) becomes a random variable. Flow models learn a neural network to estimate the expected velocity: $\bar{v}_t = \mathbb{E}[v_t \mid x_t]$. The model is trained by minimizing the following loss function, which regresses the empirical velocity for randomly sampled pairs \( (x_0, x_1) \) from the data distribution:
\begin{align}
    \mathcal{L}^{\mathrm{F}}(\theta) = \mathbb{E}_{x_0, x_1 \sim \mathcal{D}} \left[ || \bar{v}_\theta(x_t, t) - (x_1 - x_0) ||^2 \right]
    \label{eq:flow-matching}
\end{align}
To sample from the flow model, a noise point \( x_0 \) is drawn from a normal distribution and iteratively updated to \( x_1 \) via the denoising ODE, approximated using Euler sampling.

Shortcut models~\cite{shortcutmodels} further address the computational cost of flow-matching by reducing the number of sampling steps. They condition on both the timestep \( t \) and a desired step size \( d \), allowing for flexible sampling budgets and faster inference, with a self-consistency training objective.

\begin{figure}[]
\centering
  \includegraphics[width=1\columnwidth, trim={0cm 0cm 0cm 0cm}, clip]{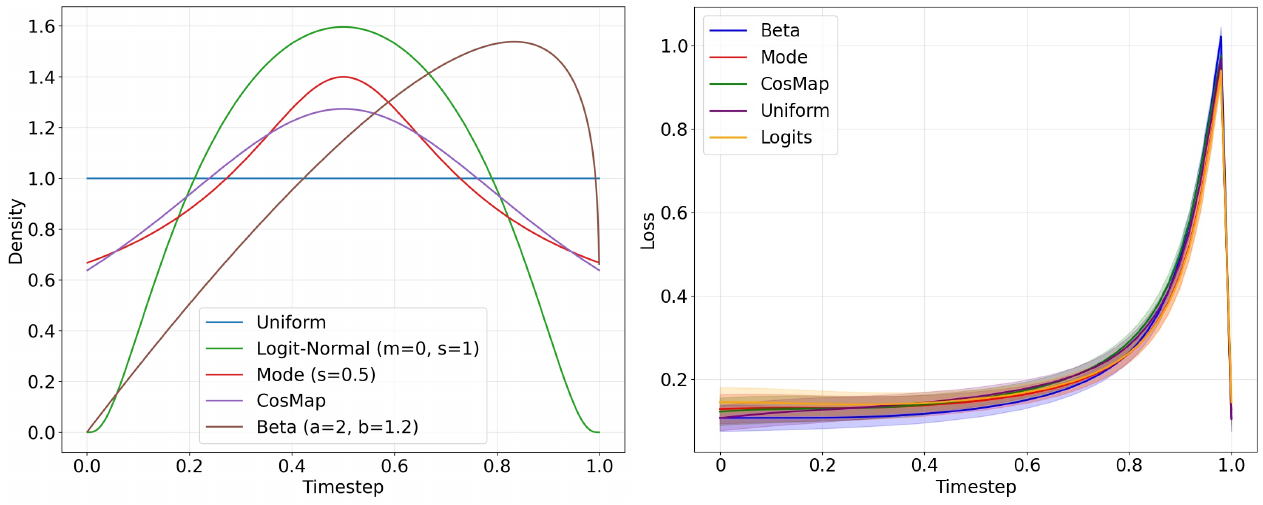}
\caption{\textbf{Time Sampling Comparison.} For various time sampling schedules, beta schedule performs the best, \textit{i.e.}, lowest training loss, with skewed pattern (\textit{left}) to counteract the ineffectiveness of model prediction when t approaches 1 (\textit{right}).
}
\label{fig:distribution}
\end{figure}

\paragraph{Gesture Latent Shortcut.}
Inspired by the principles of shortcut, we introduce GestureLSM (\underline{L}atent \underline{S}hortcut \underline{M}odel), a novel approach to address the challenges of low synthesis efficiency in gesture generation. GestureLSM achieves fast generation speeds while preserving high-quality outputs by leveraging latent shortcut learning on quantized gesture representations. 

Rather than operating directly in the image pixel space~\cite{shortcutmodels}, GestureLSM performs learning in a compact, quantized gesture motion space. As depicted in Fig.~\ref{fig:pipeline-2}, our model leverages the conditioning variable \(d\) to model future curvature, enabling the system to anticipate and directly transition to the correct target state.

The normalized direction from the current state \(x_t\) to the target state \(x'_{t+d}\) is expressed as the shortcut \(f_\theta(a, x_t, t, d)\):  
\begin{equation}
x'_{t+d} = x_t + f_\theta(a, x_t, t, d).
\label{eq:shortcut-raw}
\end{equation}
To generalize this process, we train a shortcut model \(s_\theta(x_t, t, d)\) that predicts shortcuts for all combinations of \(x_t\), \(t\), and \(d\), conditioned on the input speech signal \(a\). To ensure consistency and improve learning, we enforce a self-consistency rule:
\begin{align}
f_\theta(a, x_t, t, 2d) = \frac{1}{2} f_\theta(a, x_t, t, d) + \frac{1}{2} f_\theta(a, x'_{t+d}, t, d).
\end{align}
This rule decomposes the shortcut computation for larger step sizes into a sequence of smaller, intermediate shortcuts. By propagating this capability from multi-step processes to fewer steps and eventually to single steps, the model learns to generate high-quality gestures with minimal computational overhead.

\paragraph{Denoising Pattern Analysis.} 
Prior works have shown that time schedules significantly influence generation quality~\cite{lee2024improvingtrainingrectifiedflows,esser2024scalingrectifiedflowtransformers}, highlighting the "lost in the middle" problem in flow matching for image generation. These works propose frequent sampling in the middle timesteps to address the issue. However, we observe a distinct pattern in the audio-conditioned gesture generation domain.

As illustrated in Fig.~\ref{fig:distribution}, we evaluate various time schedules, including standard uniform sampling~\cite{liu2022flowstraightfastlearning}, Logit-Normal, Heavy-Tails, and CosMap~\cite{esser2024scalingrectifiedflowtransformers}. Across all methods, the loss consistently increases as time steps approach \( t \to 1 \), indicating poor velocity prediction at the beginning of the trajectory rather than in the middle, as seen in the image domain.

To address this, we hypothesize that a left-skewed sampling distribution can mitigate this issue. We propose using a beta distribution for timestep sampling during training:
\begin{equation}
f(t; \alpha, \beta) = \frac{t^{\alpha-1} (1-t)^{\beta-1}}{\mathrm{B}(\alpha, \beta)}, \quad t \in [0, 1],
\end{equation}
where $\alpha$, $\beta$ controls the skewness of the sampling. We compared different sampling methods and the skewness effect for model learning. From Tab.\ref{tab:ab-time}, we discover that left skewness with emphasis when t approaches 1 can significantly improve the generation quality.

\begin{figure*}[]
\centering
  \includegraphics[width=1.95\columnwidth, trim={0cm 0cm 0cm 0cm}, clip]{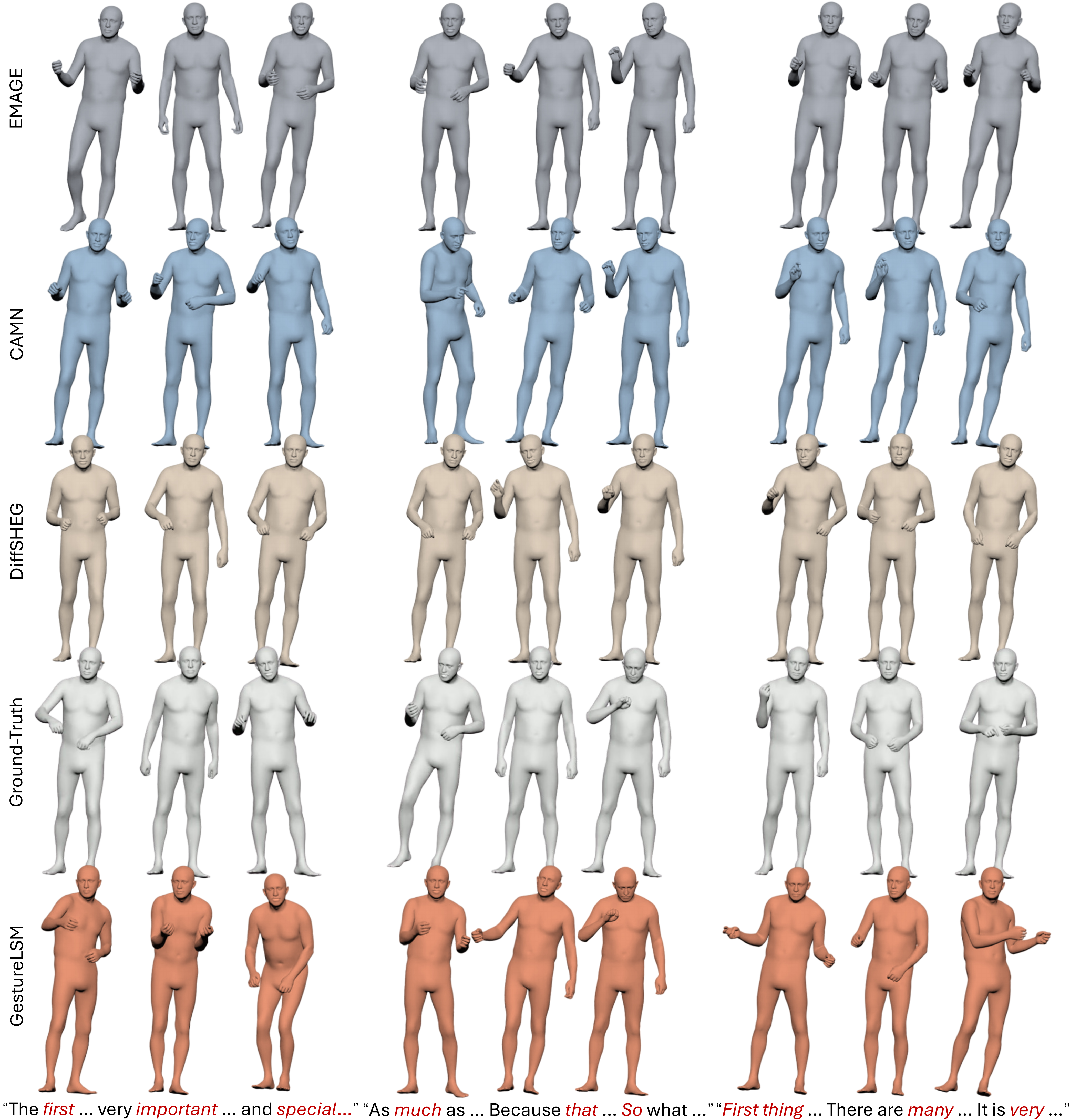}
\vspace{-0.2cm}
\caption{The subjective comparisons results. Compared with other methods, GestureLSM presents more natural gesture motions and local body region interactions. It generates more coherent full-body gestures to represent specific words or phrases (highlighted in red). }
\label{fig:visualization}
\vspace{-3mm}
\end{figure*}

\section{Experiments}
\label{sec:experiment}

\subsection{Datasets}

We train and evaluate our models using the BEAT2 dataset proposed in EMAGE~\cite{liu2023emage}. BEAT2 comprises 60 hours of high-quality SMPL-based gesture data collected from 25 speakers (12 female, 13 male). The dataset contains 1,762 sequences, each with an average duration of 65.66 seconds, where each sequence captures responses to daily inquiries. For consistency, we adopt the train-validation-test split protocol defined in EMAGE~\cite{liu2023emage}.


\subsection{Quantitative Comparisons}
\paragraph{Metrics.}
We evaluate the realism of generated body gestures using the Fréchet Gesture Distance (FGD)\cite{yoon2020speech}, which quantifies the distributional similarity between ground truth and synthesized gestures. Diversity (Div.)\cite{li2021audio2gestures} is measured by calculating the average L1 distance across multiple gesture clips. To assess speech-motion synchronization, we use Beat Constancy (BC)~\cite{li2021ai}.
For facial motion evaluation, we compute the vertex Mean Squared Error (MSE)~\cite{xing2023codetalker} to assess positional accuracy. Additionally, the efficiency of our approach is quantified through the Average Inference Time per Sentence (AITS). While previous works consider Beat Constancy higher the better, we discover the significant bias with this metric and propose to compare this metric for the generated data with that of the ground-truth data. We defer the additional details in the Appendix.

\begin{table}[ht]
    \caption{The quantitative results on BEAT. Frechet Gesture Distance (FGD) multiplied by \( 10^{-1} \), Beat Constancy (BC) multiplied by \( 10^{-1} \), Diversity, Mean Squared Error (MSE) multiplied by \( 10^{-7} \). The Average Inference Time per Sentence (AIST) is computed by one NVIDIA A100. 
    We bold the best results. }
    \label{tab:experiment}
    \centering
    \resizebox{\linewidth}{!}{
    \begin{tabular}{l|cccccc} 
    \toprule
    Methods & FGD ($\downarrow$) & BC ($\rightarrow$) & Diversity ($\uparrow$) & MSE ($\downarrow$) & AIST ($\downarrow$)\\
    \midrule
    Ground-Truth & -- & 0.703 & 11.97 & -- & --\\
    \midrule
    HA2G~\cite{liu2022learning} & 12.32 & 0.677 & 8.626  & -- & 0.195\\
    DisCo~\cite{liu2022disco}  & 9.417 & 0.643  & 9.912  & --  & 0.155\\
    $\text{CaMN}$~\cite{liu2022beat} & 6.644    & 0.676  & 10.86   & --  & 0.675\\
    $\text{DiffSHEG}$~\cite{diffsheg}  & 7.141 & 0.743 & 8.21 & 9.571  & 0.112\\
    $\text{TalkShow}$~\cite{yi2022generating} & 6.209 & 0.695 & 13.47  & 7.791  & 0.124\\
    $\text{ProbTalk}$~\cite{probtalk}  & 5.040 & 0.771 & 13.27 & 8.614 & 0.231\\
    $\text{EMAGE}$~\cite{liu2023emage} & 5.512 & 0.772 & 13.06 & 7.680 & 0.174\\
    MambaTalk~\cite{mambatalk} & 5.366 & 0.781 & 13.05 & 7.680 & 0.134\\
    $\text{SynTalker}$~\cite{chen2024syntalker} & 4.687 & 0.736 & 12.43 & -- & 2.942\\
    \midrule
    \rowcolor{mygray} GestureLSM  & \textbf{4.088} & \textbf{0.714} & 13.24 &  -- & \textbf{0.039}\\
    \rowcolor{mygray} GestureLSM (+ Face)  & \textbf{4.247} & 0.729 & \textbf{13.76} & \textbf{1.021} & 0.042\\
    \bottomrule
    \end{tabular}
    }
\vspace{-0.3cm}
\end{table}

\vspace{-0.2cm}
\paragraph{Evaluation Results.} We summarize the quantitative comparisons with existing methods in Tab.~\ref{tab:experiment}. The results highlight that our method achieves state-of-the-art performance across all evaluation metrics. We evaluate two versions of GestureLSM: one focused solely on modeling body gestures and another that integrates both body gestures and facial expressions.

Our method consistently outperforms baseline approaches, achieving a significant reduction in FGD with a score of \textbf{4.088}. This improvement stems from our effective modeling of body interactions, which minimizes unnatural gesture patterns during generation. While incorporating facial expressions slightly increases the FGD, the facial expression accuracy remains nearly indistinguishable from the ground truth, as reflected in the substantially lower Mean Squared Error (MSE) for facial expressions.

Moreover, GestureLSM achieves the the most similar BC with the ground-truth, indicating superior synchronization between speech and gestures compared to competing methods. Additionally, our model demonstrates remarkable efficiency, with an average inference speed of \textbf{0.039} seconds per frame, significantly surpassing other methods.

\begin{figure}[thb]
\centering
\includegraphics[width=\linewidth]{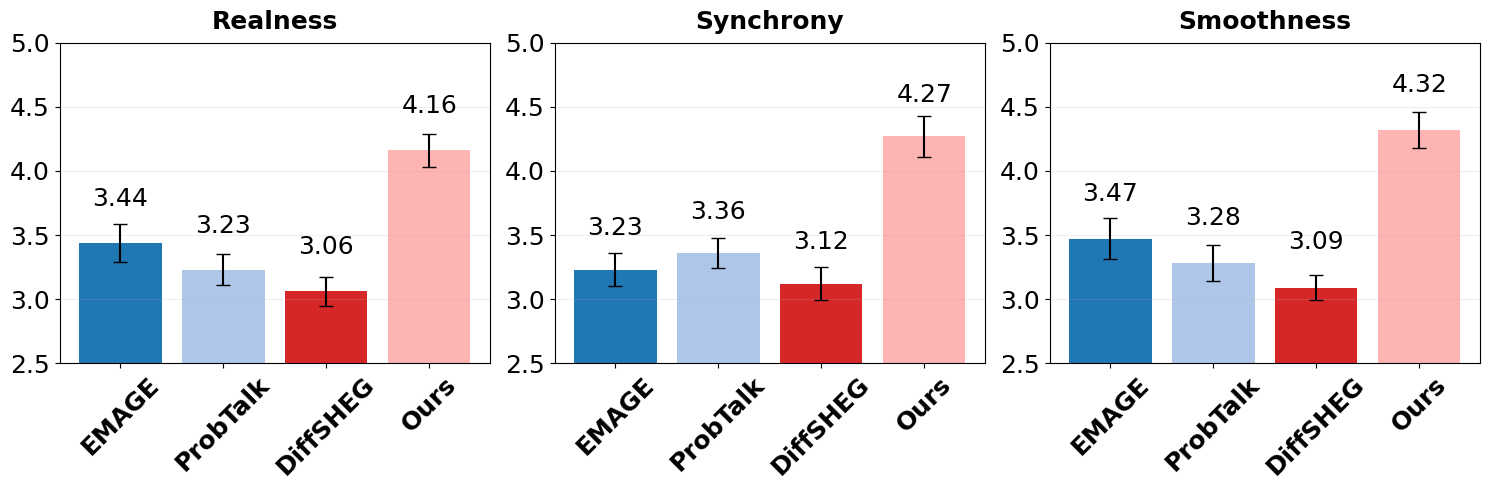}
\vspace{-0.5cm}
\caption{\textbf{User Study}. Our GestureLSM have higher user ratings with a clear margin on \textit{Realness}, \textit{Synchrony}, and \textit{Smoothness}.}
 \vspace{-0.5cm}
\label{fig:user-study}
\end{figure}

\subsection{Qualitative Comparisons}

\paragraph{Evaluation Results}
As depicted in Figure~\ref{fig:visualization}, our approach generates gestures that exhibit improved rhythmic alignment and a more natural appearance compared to existing methods. For example, when conveying the phrase \emph{``very important''}, our method directs the subject to extend both hands forward, effectively representing the intention of an emphasis. In contrast, competing methods fail to capture this nuance, often generating static or unnatural poses where one or both arms remain down.  



\vspace{-0.2cm}
\paragraph{User Study.}  
We conducted a user study with 20 participants and 320 video samples—80 from each of GestureLSM, EMAGE~\cite{liu2023emage}, ProbTalk~\cite{probtalk}, and DiffSHEG~\cite{diffsheg}—to evaluate the quality of our results. Each participant viewed the videos in a randomized order and rated them on a scale of 1 (lowest) to 5 (highest) based on three criteria: (1) \textit{realness}, (2) speech-gesture \textit{synchrony}, and (3) \textit{smoothness}.  For \textbf{realness}, participants assessed how closely the generated gestures resembled natural human movements in terms of authenticity and fluidity. For \textbf{synchrony}, they examined the timing of gestures relative to speech rhythm, audio, and facial expressions to ensure a cohesive performance.  
For \textbf{smoothness}, they analyzed motion continuity, identifying abrupt stops, unnatural jerks, and overall body coordination.  As shown in \cref{fig:user-study}, GestureLSM outperforms other methods across all criteria, achieving higher Mean Opinion Scores (MOS) and better alignment with speech by a large margin.

\begin{table*}[t]
  \caption{\small{Ablations of our method. We exam the each module contribution, model architecture design, time stamp distribution , model type analysis, speed up comparison and number of sampling steps. Bold indicates the best performance.}}
  \vspace{-0.1cm}
  \label{tab:ab}
  \renewcommand{\tabcolsep}{1.8pt}
  \small

  \begin{subtable}[!t]{0.3\linewidth}
    \centering
    \begin{tabular}{cccc}
      \toprule
      
      {\it Module.}  & FGD\(\downarrow\)  &BC$\rightarrow$   &Div.$\uparrow$  \\
    \midrule
    w/o spatial & 8.232 & 0.766 & \textbf{14.52}\\
    w/o temporal & 22.412 & 0.454 & 13.41\\
    w/o position & 4.523 & 0.656 & 14.23\\
    \rowcolor{mygray} Ours & \textbf{4.088} & \textbf{0.714} & 13.24\\
      
      \bottomrule
    \end{tabular}
    \caption{\small{Model Module Contribution.}}
    \label{tab:ab-module}
  \end{subtable}
  \hspace{\fill}
  \begin{subtable}[!t]{0.34\linewidth}
    \centering
    \begin{tabular}{cccc}
      \toprule
      
      {\it Architecture.}  & FGD\(\downarrow\)  &BC$\rightarrow$   &Div.$\uparrow$  \\
    \midrule
    Attention & 4.762 & 0.734 & 13.43 \\
    Ours + ST-Atten & 4.132 & 0.734 & 13.77 \\
    Only ST-Atten & 4.767 & 0.744 & 14.22\\
    Bi-Mamba & 4.323 & 0.745 & \textbf{14.54}\\
    \rowcolor{mygray} Ours & \textbf{4.088} & \textbf{0.714} & 13.24\\
      
      \bottomrule
    \end{tabular}
    \caption{\small{Model Architecture Design.}}
    \label{tab:ab-architecture}
    \end{subtable}
    \begin{subtable}[!t]{0.32\linewidth}
    \centering
    \begin{tabular}{cccc}
      \toprule
      
      {\it Distribution.} & FGD $\downarrow$ & BC $\rightarrow$ & Div. $\uparrow$ \\
      \midrule
    Uniform & 5.051 & 0.721 & 13.56 \\
    Logit-Normal & 4.447 & 0.755 & 13.72 \\
    Mode & 4.532 & 0.738 & 13.33 \\
    CosMap & 4.621 & 0.742 & 13.47 \\
    \rowcolor{mygray} $\alpha$=2 $\beta$=1.2 & \textbf{4.088} & 0.714 & 13.24 \\
      
    \bottomrule
    \end{tabular}
    \caption{\small{Time Stamp Sampling.}}
    \label{tab:ab-time}
    \end{subtable}

    \hspace{\fill}
    \begin{subtable}[!t]{0.3\linewidth}
    \centering
    \begin{tabular}{cccccc}
      \toprule
      {\it Model Type.} & FGD\(\downarrow\)  &Div.$\uparrow$ &AIST\(\downarrow\)\\
    \midrule
    Diffusion & 4.131 & 13.06 & 2.942 \\
    Auto-regressive & 4.232 & 13.03 & 0.153 \\
    Rectifed Flow & 4.724 & 13.21 & 0.074 \\
    Masking & 4.638 & 12.57 & 0.039 \\
    \rowcolor{mygray} ours & \textbf{4.088} & \textbf{13.24} & 0.039\\ 
      
      \bottomrule
    \end{tabular}
    \caption{\small{Generator Model Type.}}
    \label{tab:ab-type}
  \end{subtable}
  \hspace{\fill}
  \begin{subtable}[!t]{0.33\linewidth}
    \centering
    \begin{tabular}{cccccc}
      \toprule
      
      {\it Model Type.} & FGD\(\downarrow\)  &Div.$\uparrow$  &Tr-T\(\downarrow\) &AIST\(\downarrow\)\\
    \midrule
    LCM & 4.445 & 13.17 & 2.5 & 0.026 \\
    ReFlow$^{2}$ & 4.638 & 13.19 & 2 & 0.026 \\
    + Distill & 4.731 & 13.21 & 3 & 0.015 \\
    \rowcolor{mygray} Ours & \textbf{4.088} & 13.24 & \textbf{1.06} & 0.039\\ 
    \rowcolor{mygray}+ ReFlow$^{2}$ & 4.132 & \textbf{13.71} & 2.12 & \textbf{0.018}\\ 
    
      \bottomrule
    \end{tabular}
    \caption{\small{Speed Up Comparison.}}
    \label{tab:ab-speed}
  \end{subtable}
  \hspace{\fill}
    \begin{subtable}[!t]{0.33\linewidth}
    \centering
    \begin{tabular}{ccccc}
      \toprule
      {\it Steps.}  & FGD\(\downarrow\)  &BC$\rightarrow$  &Div.$\uparrow$ & AIST\(\downarrow\)\\
      \midrule
      1 & 6.235 & 0.647 & 13.23 & \textbf{0.015} \\
      2 & 4.988 & 0.680 & 13.39 & 0.018\\ 
      4 & 4.262 & \textbf{0.704} & \textbf{13.35} & 0.026\\ 
      \rowcolor{mygray} 8 & 4.088 & 0.714 & \textbf{13.24} & 0.039\\ 
      20 & \textbf{4.040} & 0.730 & 13.49 & 0.076\\ %
      \bottomrule
    \end{tabular}
    \caption{\small{Number of sampling steps.}}
    \label{tab:ab-sample}
  \end{subtable}


    

  \vspace{-5mm}
\end{table*}

\subsection{Ablation Studies}

To verify our designs, we conduct ablations. we defer additional ablation studies in the Appendix.

\vspace{-0.2cm}
\paragraph{Model Module Contribution.}
We analyze each module contribution: (1) w/o spatial: Only temporal attention. (2) w/o temporal: Only spatial attention. (3) w/o position: Remove spatial and temporal positional encoding. Tab.~\ref{tab:ab-module} shows applying both spatial temporal attentions achieves the best results, with positional encoding providing gains.

\vspace{-0.2cm}
\paragraph{Model Architecture Design.}
We analyze architectural variations: (1) Attention: Replace spatial and temporal attentions with standard attention~\cite{vaswani2023attentionneed}.  (2) Ous + ST-Atten: Adding a spatial-temporal attention layer after spatial and temporal attention by flattening features into a 1D sequence, similar to ViT~\cite{dosovitskiy2021imageworth16x16words}. (3) Only ST-Atten: Removing separate spatial and temporal attentions, using only spatial-temporal attention. (4) Bi-Mamba: Adopting MambaTalk's~\cite{mambatalk} global-local scanning for the generator architecture. Tab.~\ref{tab:ab-architecture} shows spatial-temporal attention alone does not improve interaction learning, likely due to the complexity of flattening features. Neither standard attention nor Bi-Mamba surpasses our model, validating our design.

\vspace{-0.2cm}
\paragraph{Time Stamp Sampling.}
We evaluate different time sampling stamps: As shown in Tab.~\ref{tab:ab-time}, uniform is the standard training method for rectified flow models. The middle three rows are sampling methods utilized in Stable Diffusion 3~\cite{esser2024scalingrectifiedflowtransformers} with an emphasis in the middle. The last row represents the proposed beta-schedule. It can be seen the left-skewed time stamp sampling can effectively improve the gesture generation quality.



\vspace{-0.2cm}
\paragraph{Model Type.}
To isolate the impact of model type and design, we compare: (1) Diffusion: Use DDPM~\cite{ho2020denoisingdiffusionprobabilisticmodels}. (2) Auto-regressive: Next-token prediction following EMAGE~\cite{liu2023emage} and MambaTalk~\cite{mambatalk}. (3) Rectified Flow: Standard rectified flow~\cite{liu2022flowstraightfastlearning} with uniform time sampling. (4) Masking: We adopt masking-based generation from Momask~\cite{momask} for training and iterative remasking during inference. Tab.~\ref{tab:ab-type} shows other model types achieve slight inferior performance with slow inference speed.

\vspace{-0.2cm}
\paragraph{Speed Up Comparison.}
To verify the generation speed and quality, we further compare the existing speed-up methods for various model types: (1) Diffusion+LCM: Similar to MotionLCM~\cite{motionlcm}, we add latent consistency distillation for the base Diffusion. (2) ReFlow$^{2}$: We fine-tune the Rectified Flow base model by Reflow based on ~\cite{liu2022flowstraightfastlearning}. (3) ReFlow$^{2}$ + Distill: We further distill the ReFlow model into one-step model. To validate the training efficiency, we let the training time of base models from Tab.~\ref{tab:ab-type} as one unit time. Tab.~\ref{tab:ab-speed} shows Latent consistency distillation fails to match our method's training efficiency or performance. ReFlow with Distillation is under-performed. Our latent shortcut outperforms all these works with only 1.06 unit of training time. Note that we can further apply ReFlow$^{2}$ to our shortcut model and achieve one or two-step sampling generation with slight performance drop.

\vspace{-0.2cm}
\paragraph{Sampling Steps.}
Performance with different sampling steps is shown in Tab.~\ref{tab:ab-sample}. Even with two steps, the model achieves an FGD of 4.988 with an inference time of 0.018, already outperforming prior methods. Additional steps refine performance further. We take eight steps for inference efficiency and high generation quality.

\subsection{Application}
\begin{figure}[t]
    \centering
    \includegraphics[width=1\linewidth]{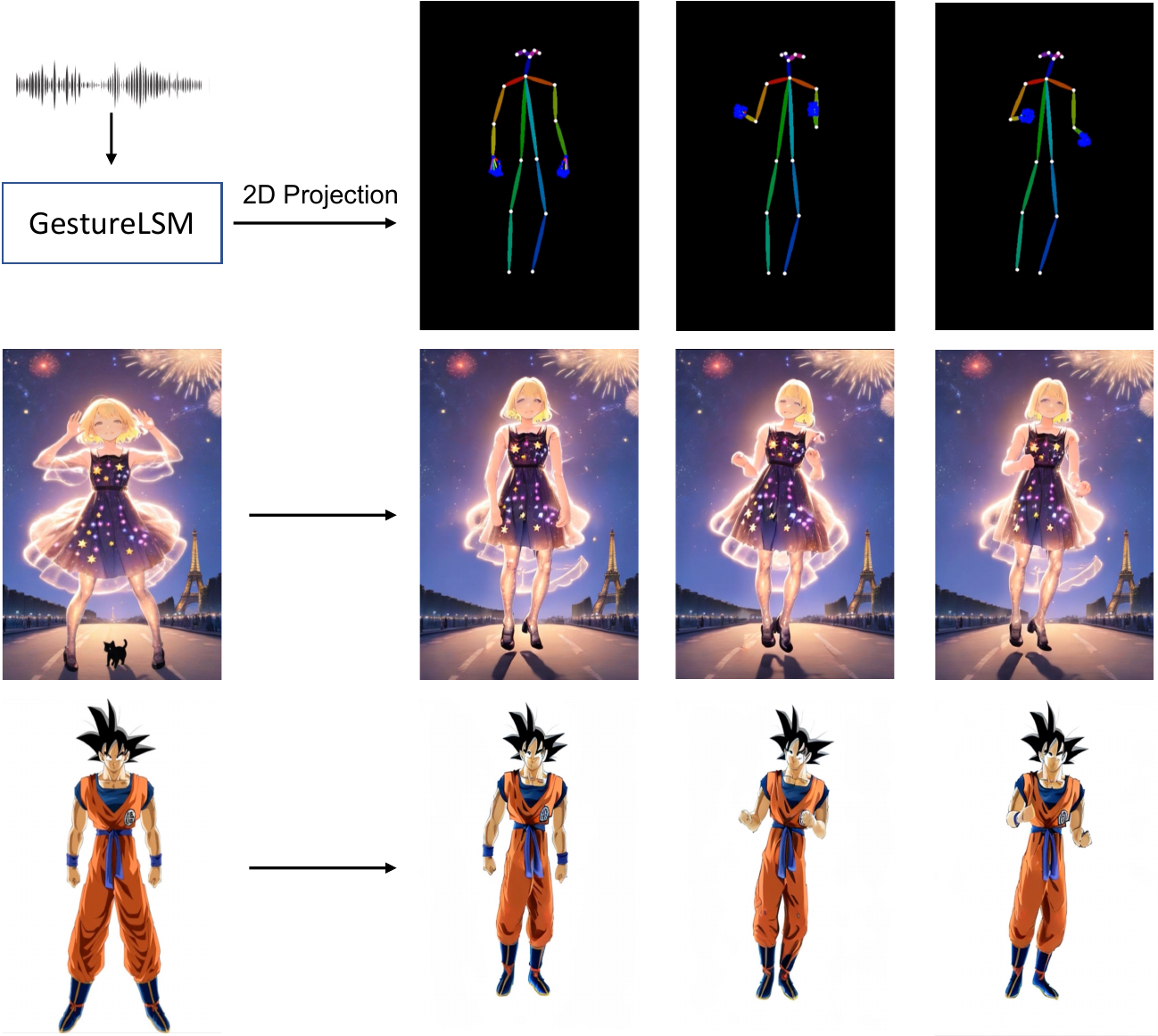}
    \caption{\textbf{Avatar Video Generation Application.} After generating a sequence of gesture motions conditioned on speech audio by GestureLSM, we project the 3D keypoints to a 2D plane, serving as keypoint guidance for avatar video generation based on 2D animation methods like AnimateAnyone~\cite{hu2023animateanyone}}
    \label{fig:animate}
    \vspace{-0.4cm}
\end{figure}

In Fig.~\ref{fig:animate}, we show two videos that are generated based on our synthesized gesture motion sequences. As a vital application direction, users can leverage GestureLSM to first generate the 3D-SMPL~\cite{SMPL-X:2019} poses and projected to 2D spaces as gesture keypoints for customized avatar animations with the assistance of off-the-shelf motion-guided video generation technology like AnimateAnyone~\cite{hu2023animateanyone}, users can freely create videos for their favorite characters.
\section{Conclusion}
We present \textbf{GestureLSM}, a latent shortcut framework that achieves real time speech driven gesture generation. Our method explicitly represent the gesture into different body regions and leverage spatial and temporal attentions to model their interactions. To leverage shortcut model for the gesture generation framework, we further propose an effective beta distribution for time stamp during training. Extensive comparisons show that \textbf{GestureLSM} improves co-speech gesture generation and achieves real-time inference for various downstream applications. 

{
    \small
    \bibliographystyle{ieeenat_fullname}
    \bibliography{main}
}

\clearpage
\appendix

\section{Overview}
\label{sec:Summary}
The supplementary document contains implementation details, metric details, additional experimental results and training analysis. For more visual results, \textbf{please see the demo videos}.

\section{Implementation Details}
In the construction of the RVQVAEs, the codebook is initialized uniformly, with each entry having a feature length of 128 and a total size of 1,024 per body region. The codebook updates occur solely during the quantization process, with resets following Contextual Gesture~\cite{liu2025contextualgesturecospeechgesture}. The RVQVAEs are trained for 30,000 iterations, with a learning rate of \( 2 \times 10^{-4} \). The GestureLSM model contains 3 layers of cross-attention for audio-gesture feature fusion and 8 layers of spatial-temporal attention blocks. The latent dimension is set to 256 with feed-forward size of 1024. During the second training stage for speech-to-gesture generation, the codebook remains frozen. We train the GestureLSM model for 1000 epochs. We utilize the Adam optimizer with a learning rate of \( 2 \times 10^{-4} \). All experiments are conducted on a single NVIDIA A100 GPU. We adopt a guidance dropout rate of 0.1 during training and a speech-conditioning ratio of 2 during generation.

\section{Metric Details}

\paragraph{Fr\'{e}chet Gesture Distance (FGD)}
Fr\'{e}chet Gesture Distance (FGD), introduced in \cite{yoon2020speech}, quantifies the similarity between the distributions of real and generated gestures, where a lower FGD signifies a closer match. Inspired by perceptual loss in image generation, FGD is computed using latent features extracted from a pretrained network:
\begin{equation}
\label{eqfid}
\resizebox{.85\hsize}{!}{$
\operatorname{FGD}(\mathbf{g}, \hat{\mathbf{g}})=\left\|\mu_{r}-\mu_{g}\right\|^{2}+\operatorname{Tr}\left(\Sigma_{r}+\Sigma_{g}-2\left(\Sigma_{r} \Sigma_{g}\right)^{1 / 2}\right),$}
\end{equation}
where $\mu_{r}$ and $\Sigma_{r}$ denote the mean and covariance of the latent feature distribution $z_{r}$ derived from real gestures $\mathbf{g}$, while $\mu_{g}$ and $\Sigma_{g}$ correspond to the statistics of the generated gestures $\hat{\mathbf{g}}$.

\paragraph{L1 Diversity}
L1 Diversity, proposed in \cite{li2021audio2gestures}, measures the variation across multiple gesture sequences, with higher values indicating greater diversity. The average L1 distance across $N$ motion sequences is computed as:
\begin{equation}
\resizebox{.75\hsize}{!}{$
    \text{L1 div.} =  \frac{1}{2 N (N-1)} \sum_{t=1}^{N} \sum_{j=1}^{N} \left\|p_{t}^{i}-\hat{p}_{t}^{j}\right\|_{1},$}
\end{equation}
where $p_{t}$ represents the joint positions at frame $t$. Diversity is evaluated on the complete test set. To ensure a focus on local motion, global translation is neutralized when computing joint positions.

\paragraph{Beat Constancy (BC)}
Beat Constancy (BC), as defined in \cite{li2021ai}, assesses the temporal alignment between gestures and audio rhythm. Higher BC values indicate stronger synchronization. Speech onsets are treated as audio beats, while motion beats correspond to local minima in the upper body joint velocity (excluding fingers). The alignment is determined using:
\begin{equation}
\label{align}
\resizebox{.80\hsize}{!}{$
\text{BC}= \frac{1}{g} \sum_{b_{g}\in g} \exp \left(-\frac{\min _{b_{a}\in a}\left\|b_{g}-b_{a}\right\|^{2}}{2 \sigma^{2}}\right),$}
\end{equation}
where $g$ and $a$ denote the sets of detected gesture beats and audio beats, respectively.

\section{Additional Experimental Results}
\label{sec:ab-additional}

\vspace{-0.1cm}
\paragraph{Feature Contributions.}
We assess feature variations: (1) w/o text: Exclude speech transcripts. (2) wavLM: Replace the CNN audio encoder with pretrained WavLM~\cite{chen2022wavlm}. (3) concatenate: Use concatenation with an MLP for fusion instead of cross-attention. (4) addition: Element-wise addition of speech and gesture features. Tab.~\ref{tab:ab-feature} shows cross-attention consistently outperforms other fusion methods, while WavLM provides no advantage.

\vspace{-0.1cm}
\paragraph{Classifier Free Guidance.}
We evaluate guidance scale for conditional generation. We show their performance by the same number of sampling steps of 8. Tab.\ref{tab:ab-cfg} shows a guidance scale of 2 achieves the best performance.

\vspace{-0.1cm}
\paragraph{Gesture Representation.}
We evaluate gesture quantization methods: (1) w/o quant: Directly use 6D-rotations of joints, (2) one quant: Single VQ quantizer for the whole body. (3) one residual: Single RVQ quantizer for the whole body. (4) product quant: 2D quantizer based on ProbTalk~\cite{probtalk}. Tab.~\ref{tab:ab-represent} shows RVQ outperforms VQ and product quantization. Separating body regions further improves performance over holistic representations.

\vspace{-0.1cm}
\paragraph{Sequential Design of Attention.}
We analyze the sequential design of the proposed two types od attentions. Tab.\ref{tab:ab-seq} shows attention in spatial-temporal order present slightly improvement.

\vspace{-0.1cm}
\paragraph{Skewness of Time Distribution.}
We further evaluate the skewness of the proposed beta schedule for time stamp distirbution. Tab.~\ref{tab:ab-skew} shows with $\beta = 1.2$ and $\alpha = 2.0$ achieves the best performance. This indicate the emphasize and a more significant left skewness with an emphasis appraoching to 1 is important for the model learning.

\begin{figure*}[t]
    \centering
    \begin{subfigure}{0.32\textwidth}
        \centering
        \includegraphics[width=\linewidth]{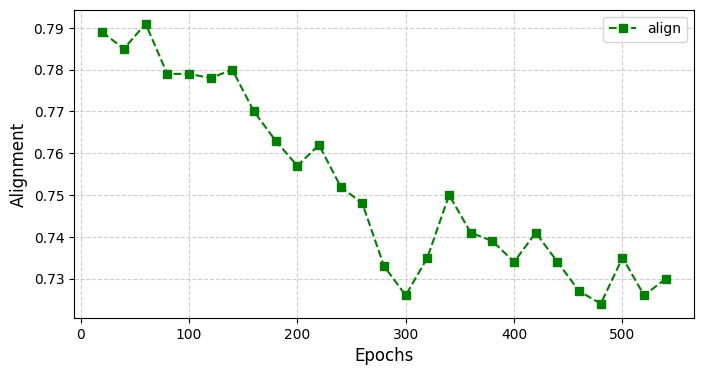}
        \caption{Beat constancy over training.}
        \label{fig:ab-align}
    \end{subfigure}
    \hfill
    \begin{subfigure}{0.32\textwidth}
        \centering
        \includegraphics[width=\linewidth]{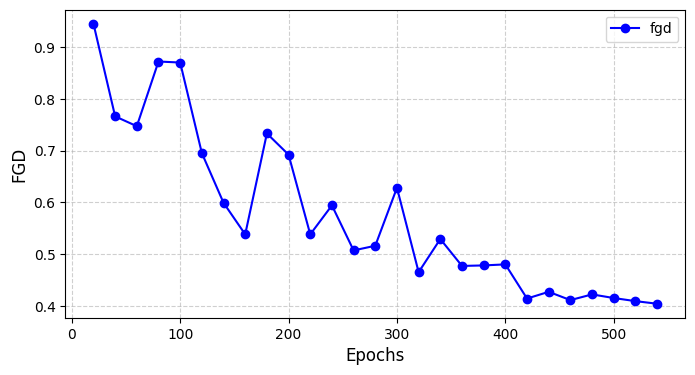}
        \caption{Fréchet Gesture Distance (FGD) over training. }
        \label{fig:ab-fgd}
    \end{subfigure}
    \hfill
    \begin{subfigure}{0.32\textwidth}
        \centering
        \includegraphics[width=\linewidth]{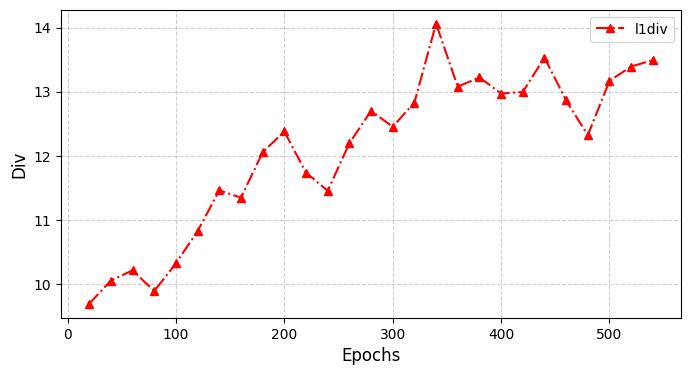}
        \caption{Gesture diversity over training.}
        \label{fig:ab-div}
    \end{subfigure}
    
    \vspace{-0.3cm}
    \caption{Training dynamics of key evaluation metrics. (a) Beat constancy decreases, indicating a shift from overly rigid beat-following motions to more natural gestures. (b) FGD decreases, reflecting improved gesture realism. (c) Gesture diversity increases, suggesting a broader range of motion patterns learned by the model.}
    \label{fig:ab-metrics}
    \vspace{-0.3cm}
\end{figure*}

\begin{table*}
  \caption{\small{Additional ablations of our method. We exam the speech feature,  classifier free guidance scale, gesture representation, sequence order for the attention and the skewness for the sampling distribution. Bold indicates the best performance.}}
  \vspace{-0.1cm}
  \label{tab:ab}
  \renewcommand{\tabcolsep}{1.8pt}
  \small

\begin{subtable}[!t]{0.3\linewidth}
    \centering
    \begin{tabular}{ccccc}
      \toprule
      {\it Features.} & FGD\(\downarrow\)  &BC$\rightarrow$  &Div.$\uparrow$\\
      \midrule
      w/o text & 4.323 & 0.743 & 13.17 \\
      w WavLM & 4.567 & \textbf{0.707} & 13.23 \\
      concatenate & 4.676 & 5.479 & 11.67 \\
      addition & 6.012 & 6.234 & 13.11 \\
      \rowcolor{mygray} cross-atten & \textbf{4.088} & 0.714 & \textbf{13.24} \\
      \bottomrule
    \end{tabular}
    \caption{\small{Speech Feature.}}
    \label{tab:ab-feature}
  \end{subtable}
  \begin{subtable}[!t]{0.3\linewidth}
    \centering
    \begin{tabular}{cccc}
      \toprule
      {\it Scale.} & FGD\(\downarrow\)  &BC\(\rightarrow\)  &Div.\(\uparrow\)\\
      \midrule
      1.0 & 4.215 & 0.741 & 12.79 \\
      1.5 & 4.141 & 0.730 & 13.26 \\
      \rowcolor{mygray} 2.0 & \textbf{4.088} & 0.714 & 13.24 \\
      2.5 & 4.124 & 0.714 & 13.61 \\
      3.0 & 4.157 & \textbf{0.709} & \textbf{13.75}\\
      \bottomrule
    \end{tabular}
    \caption{\small{CFG Scale.}}
    \label{tab:ab-cfg}
  \end{subtable}
  \begin{subtable}[!t]{0.3\linewidth}
    \centering
    \begin{tabular}{cccc}
      \toprule
      {\it Represent.}     & FGD\(\downarrow\)  & BC$\rightarrow$ & Div.$\uparrow$\\
    \midrule
    w/o quant & 8.727 & 0.612 & \textbf{13.56} \\
    one quant & 6.343 & 0.734 & 13.42 \\
    one residual & 5.256 & 0.755 & 13.35 \\
    product quant & 4.412 & 0.737 & 13.41 \\
    \rowcolor{mygray} Ours & \textbf{4.088} & \textbf{0.714} & 13.24\\
      \bottomrule
    \end{tabular}
    \caption{\small{Gesture Motion Representation.}}
    \label{tab:ab-represent}
  \end{subtable}

  \begin{subtable}[!t]{0.3\linewidth}
    \centering
    \begin{tabular}{cccc}
      \toprule
      {\it Order.} & FGD\(\downarrow\)  &BC\(\rightarrow\)  &Div.\(\uparrow\)\\
      \midrule
      \rowcolor{mygray} spatial-temporal & \textbf{4.088} & \textbf{0.714} & 13.24 \\
      temporal-spatial & 4.113 & 0.721 & \textbf{13.34} \\
      \bottomrule
    \end{tabular}
    \caption{\small{Sequence Order.}}
    \label{tab:ab-seq}
  \end{subtable}
  \begin{subtable}[!t]{0.3\linewidth}
    \centering
    \begin{tabular}{lccc}
    \toprule
    {\it Distribution.} & FGD $\downarrow$ & BC $\rightarrow$ & Div. $\uparrow$ \\
    \midrule
    \rowcolor{mygray} $\alpha$=2 $\beta$=1.2 & \textbf{4.088} & 0.714 & 13.24 \\
    $\alpha$=2 $\beta$=1.0 & 4.123 & \textbf{0.704} & 13.44 \\
    $\alpha$=2.2 $\beta$=1.4 & 4.362 & 0.754 & 13.65 \\
    $\alpha$=1.8 $\beta$=1.4 & 4.341 & 0.743 & \textbf{13.73} \\
    \bottomrule
    
    \end{tabular}
    \caption{\small{Skewness of the Distribution.}}
    \label{tab:ab-skew}
  \end{subtable}
  
\end{table*}

\section{Training Analysis}
\label{sec:ab-train}
We analyze how the model's performance evolves during training. As shown in Fig.~\ref{fig:ab-align}, the model exhibits the highest beat constancy at the early stages of training. However, we observe that this corresponds to unnatural, exaggerated motion patterns in response to speech beats. As training progresses, beat constancy gradually decreases. Importantly, we argue that higher beat constancy is not necessarily better. For reference, the ground-truth gestures exhibit a beat constancy of 0.703, suggesting that aligning this metric with real human motion is a more meaningful target. Based on this, we propose evaluating beat constancy relative to the ground-truth rather than treating higher values as strictly superior.

For other metrics, we observe a clear upward trend in gesture diversity and a corresponding decrease in FGD as training progresses. These trends align with existing literature, and we maintain the standard evaluation approach for these metrics.


\end{document}